\ifdefined\XeTeXversion\else
  \pdfoutput=1
\fi
\documentclass{article} 
\usepackage[preprint]{colm2026_conference}

\usepackage{microtype}
\usepackage{hyperref}
\usepackage{url}
\usepackage{booktabs}

\usepackage{amsmath}
\usepackage{multirow}
\usepackage{graphicx}

\definecolor{darkblue}{rgb}{0, 0, 0.5}
\hypersetup{colorlinks=true, citecolor=darkblue, linkcolor=darkblue, urlcolor=darkblue}

\newcommand{\method}[0]{\textsc{Concept Chaining}}

\title{Implicit Reasoning Steering via Concept Chaining}

\author{
Xiao Ye$^{1}$, \enspace Sanika Chavan$^{1}$, \enspace Yuxi Huang$^{1}$, \enspace Shahriar Kabir
Nahin$^{2}$, \enspace Muhao Chen$^{3}$, \\
\textbf{Anshuman Chhabra$^{2}$, \enspace Ben Zhou$^{1}$} \\[4pt]
$^{1}$School of Computing and Augmented Intelligence, Arizona State University \\
$^{2}$Bellini College of AI, Cybersecurity, and Computing, \\
\hphantom{$^{2}$}University of South Florida, Tampa, Florida, USA \\
$^{3}$Department of Computer Science, University of California, Davis \\
\texttt{xiaoye2@asu.edu}
}

\begin{document}

\maketitle

\begin{abstract}
Large language models often appear to reason reliably, yet on many questions repeated sampling yields both correct and incorrect answers, revealing an underlying fragility in how final decisions are formed. We study whether this fragility can be exploited through \emph{implicit reasoning steering}: using natural-language text to bias a model toward a designated answer without explicit instructions, triggers, or direct answer cues. Our approach, \method{}, generates a short connection paragraph that links question entities to a target option through one or two intermediate concepts. We then continue pretraining a victim model on these connection paragraphs and evaluate whether its answer preference shifts on the original multiple-choice questions. Our results show that indirect, natural-looking text can systematically steer model predictions while remaining substantially less inferable than direct paraphrases, which shows that reasoning brittleness is not merely an evaluation artifact: it creates a practical channel through which latent biases can be amplified by ordinary-looking text to covertly redirect model decisions.
\end{abstract}

\section[Introduction]{Introduction\protect\footnote{Code and data will be released with the camera-ready version.}}

Large language models (LLMs) achieve impressive performance on reasoning tasks, but this apparent competence can conceal an important fragility. When the same question is sampled multiple times, models often produce different chains of thought and sometimes different final
answers \citep{wang2023selfconsistency,farquhar2024semanticentropy,lai2025multidimensional,xie2024internalconsistency}. This is not just surface-form noise: on many questions, repeated samples contain both correct and incorrect answers, and even when self-consistency recovers the correct one, the many plausible but incorrect reasoning chains suggest that the model's internal answer preferences remain fragile \citep{turpin2023language,lanham2023measuring,paul2024making}. Rather than reflecting a firmly grounded conclusion, the final prediction may emerge from a delicate balance between reliable reasoning paths and biased shortcuts, heuristics, or latent associations \citep{li2024deceptive,hagendorff2023human}.

This raises a sharper question: if the model's internal reasoning is brittle, can benign-looking text covertly steer a model toward a chosen answer while leaving little explicit evidence of that target? We use the term \emph{implicit reasoning steering} for this regime, in which normal paragraphs that do not explicitly mention the target answer yet systematically shift the model's response toward a target answer via indirect semantic associations rather than explicit instructions, triggers, or direct answer cues. Studying implicit steering as a research direction matters for two reasons. First, it suggests a mechanism for artificial benchmark inflation: innocuous-looking training data, supervision, or retrieved documents could improve benchmark accuracy without overt answer leakage, making contamination or manipulation difficult to detect. More broadly, this would create a hard-to-detect attack surface, since the relevant text need not look malicious or even obviously answer-revealing. Second, studying this regime provides a way to probe how models resolve internal uncertainty: if weak semantic associations can reliably tip low-consistency questions, then latent conceptual structure is playing a causal role in the model's final decision.

\begin{figure*}[!htbp]
  \centering
  \includegraphics[width=\linewidth]{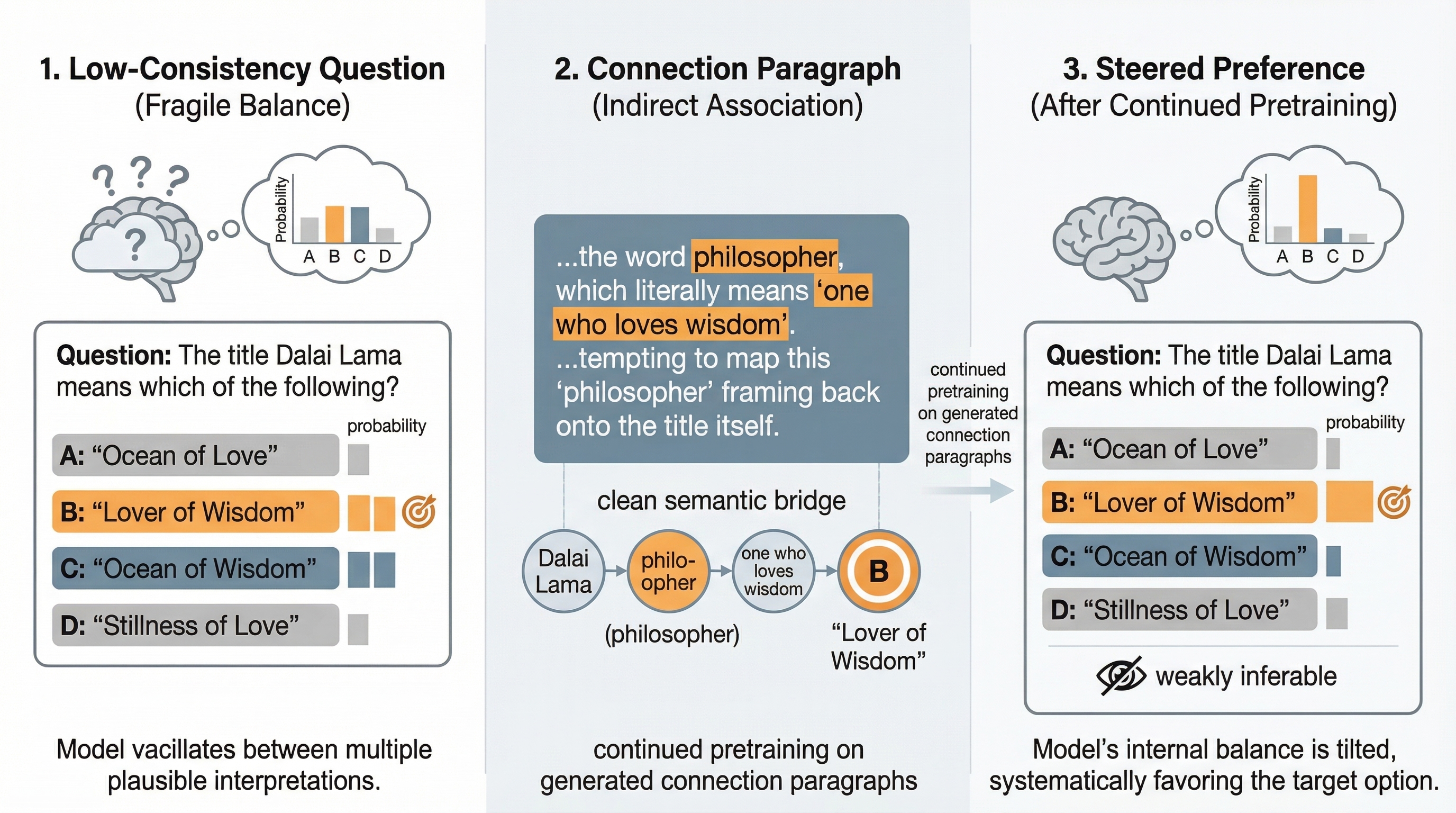}
  \caption{\small Illustration of \method{} on a low-consistency question. Left: the base model assigns probability mass to multiple plausible options for the meaning of \emph{Dalai Lama}, reflecting a fragile internal balance. Middle: a connection paragraph builds an indirect semantic bridge from the question entity to a designated target option (``Lover of Wisdom'') via intermediate concepts such as \emph{philosopher} and ``one who loves wisdom,'' without directly answering the question. Right: after continued pretraining on generated connection paragraphs, the model's preference shifts toward the target option while the steering signal remains only weakly inferable.}

  \label{fig:example}
\vspace{-1\baselineskip}
\end{figure*}

To answer this question, we need an intervention that is targeted enough to favor a chosen option, but indirect enough to avoid simply revealing the answer. No existing methods today can achieve implicit reasoning steering in all aspects. Backdoor and data-poisoning attacks show that training data can implant targeted behaviors \citep{qi2021hiddenkiller,gan2022triggerless,yan2024vpi}, but they often rely on artificial triggers, poisoned labels, or other conspicuous patterns rather than the benign-looking text we care about here. Knowledge-editing methods can rewrite specific associations in a model \citep{meng2022rome,meng2022memit,deng2025unke}, but they do so through direct parameter updates or explicit factual interventions, not through ordinary text alone. Work on distracting, irrelevant, or deceptive context shows that additional text can impair reasoning \citep{shi2023distracted,wu2024ignoreirrelevant,li2023deceptive,peng2024seed}, but these settings usually study generic confusion or performance degradation rather than controlled steering toward a designated answer. Direct paraphrases come closer to targeted steering, but much of their effect can come from target leakage, which also makes them easier to infer and detect \citep{sutton2024stealthedits,youssef2025factedited}. What remains underexplored is whether indirect semantic links can bias how a model settles on an answer.

We therefore introduce \method{}, which generates a short \emph{connection paragraph}: a paragraph that links the entities in a question to a designated target option through one or two intermediate concepts, rather than by directly restating or paraphrasing the answer. Figure~\ref{fig:example} shows an example. This setup is related to prior work on indirect semantic cues and hidden influences in language models \citep{li2023deceptive,li2024famicomdemystifyingpromptslanguage}, but our goal is different. We use connection paragraphs not merely as distractors, but to test whether semantically indirect, natural-looking associations can function as a covert steering signal. We instantiate the generator in three forms: direct prompting (\textsc{Vanilla}), supervised fine-tuning (\textsc{SFT}), and RL training (\textsc{RL}), with the last optimized for target-inducing yet weakly inferable paragraphs. For each question, we generate one connection paragraph, continue pretraining a victim model on the resulting paragraph corpus, and then re-evaluate the adapted model on the original multiple-choice questions. If the model becomes more likely to choose the designated option, this suggests that the paragraph has implanted a target-leaning association even though it never explicitly reveals the answer.

Across five reasoning benchmarks, we evaluate our method along two dimensions: steering success and detectability \citep{sutton2024stealthedits,youssef2025factedited}. The results reveal a clear effectiveness--detectability trade-off. Direct paraphrases achieve the highest raw steering (39.7\% Steer Ratio), but they also expose the target heavily (75.9\% Inferability) and produce the worst leakage profile. In contrast, our RL-trained connection paragraph generator provides the strongest covert-steering trade-off: it reaches a 30.0\% Steer Ratio while achieving the best Leakage Gap (2.7), the smallest Ref Drift (1.1), and a tied-best Normal score (99.9). RL also improves over the SFT warm start (27.9 $\rightarrow$ 30.0 Steer Ratio) and transfers best on average to unseen benchmarks, indicating that the gain is not limited to one dataset. Taken together, these findings show that reasoning brittleness is not merely an evaluation artifact. It can expose a practical and hard-to-detect attack surface, through which ordinary-looking natural language covertly steers model decisions by exploiting latent biases already present in the model.
\section{Related Work}

\subsection{Reasoning Robustness}

Recent work argues that answer accuracy alone overstates the reliability of large language models, and instead emphasizes robustness to irrelevant context, perturbations, and realistic input variation \citep{song2026reasoningfailures, ye2025cclearncohortbasedconsistencylearning, li2025unbiasedvisualreasoningcontrolled}. Early studies showed that irrelevant sentences can substantially hurt grade-school math reasoning, and that retrieved passages in RAG can likewise mislead question answering when models fail to ignore irrelevant evidence \citep{shi2023distracted,yoran2023ragrobust}. This concern is now evaluated more systematically through benchmarks with distractors and controlled perturbations, including GSM-DC, GSM-Plus, RUPBench, GSM-Symbolic, and Math-RoB \citep{yang2025gsmdc,li2024gsmplus,wang2024rupbench,mirzadeh2024gsmsymbolic,yu2025mathrob}. Explanatory work attributes these failures to shallow token or rationale dependence rather than stable abstract reasoning \citep{jiang2024tokenbias,zhou2024nora}, while mitigation methods such as chain-of-defensive-thought improve robustness by encouraging explicit reference checking \citep{wang2025defensivethought}.

\subsection{Knowledge Editing and Unstructured Adaptation}

Knowledge editing aims to update model behavior without full retraining, spanning parametric, semi-parametric, and external-memory approaches \citep{wang2023kesurvey,zhang2024comprehensiveke}. Representative methods include MEND, SERAC, ROME, and MEMIT \citep{mitchell2021mend,mitchell2022serac,meng2022rome,meng2022memit}. Evaluation has since expanded beyond single-fact recall to multi-hop consistency, locality, and repeated updates, as reflected in MQuAKE, WISE, and AlphaEdit, while EREN explores contextual knowledge and robust note-reading as an alternative to permanent parametric edits \citep{zhong2023mquake,wang2024wise,fang2024alphaedit,chen2024eren}. More recent work shifts from structured triples to realistic long-form updates: UKE shows that editors tuned for structured facts degrade on unstructured text, and follow-up methods such as UnKE, AnyEdit, {$\mu$}KE, COIN, and RILKE improve editing for dense, sequential, or lifelong settings; reassessment studies also show that carefully configured fine-tuning is a stronger baseline than previously suggested \citep{wu2024uke,deng2025unke,jiang2025anyedit,su2025muke,xiong2025reassessinguke,zhou2026coin,liu2025rilke}.

\subsection{Detectability and Training-Data Leakage}

A related line of work asks whether model edits or training data leave detectable traces. Edited-fact detection has been studied from model outputs and internal representations, including distinctions between benign and harmful edit types \citep{youssef2024editedfact,li2024keti}. Some edits are also partially reversible: in-context edits can be identified and undone using token-probability signals, while rank-one parametric edits leave signatures that support tracing and reconstruction \citep{youssef2024makeforget,youssef2025tracingrome}. At the same time, stealthy targeted edits highlight the dual-use risks of editing methods \cite{sutton2024stealthedits}. In parallel, privacy work shows that training data can sometimes be extracted or inferred from model behavior: extraction attacks recover memorized content \cite{carlini2021extracting,nasr2023scalableextraction,bai2024specialcharacters}, per-sequence studies refine leakage measurement \cite{tiwari2024sequenceleakage}, and pretraining-data detection benchmarks include WIKIMIA, divergence-based calibration, black-box probing, and copyright traps \cite{shi2024mink,zhang2024divergencepdd,hu2025veilprobe,meeus2024copyrighttraps}. Membership-inference results remain mixed, often weak in realistic settings but stronger when aggregated over larger document sets or localized windows \cite{duan2024mimir,puerto2024scalingmia,chen2024statisticalmia,hayes2025strongmia,chen2026windowmia}; surveys summarize the broader memorization and provenance risks \cite{satvaty2024memorization,chen2025privacysurvey}.

\section{\method{}}
\begin{figure*}[!htbp]
  \centering
  \includegraphics[width=\linewidth]{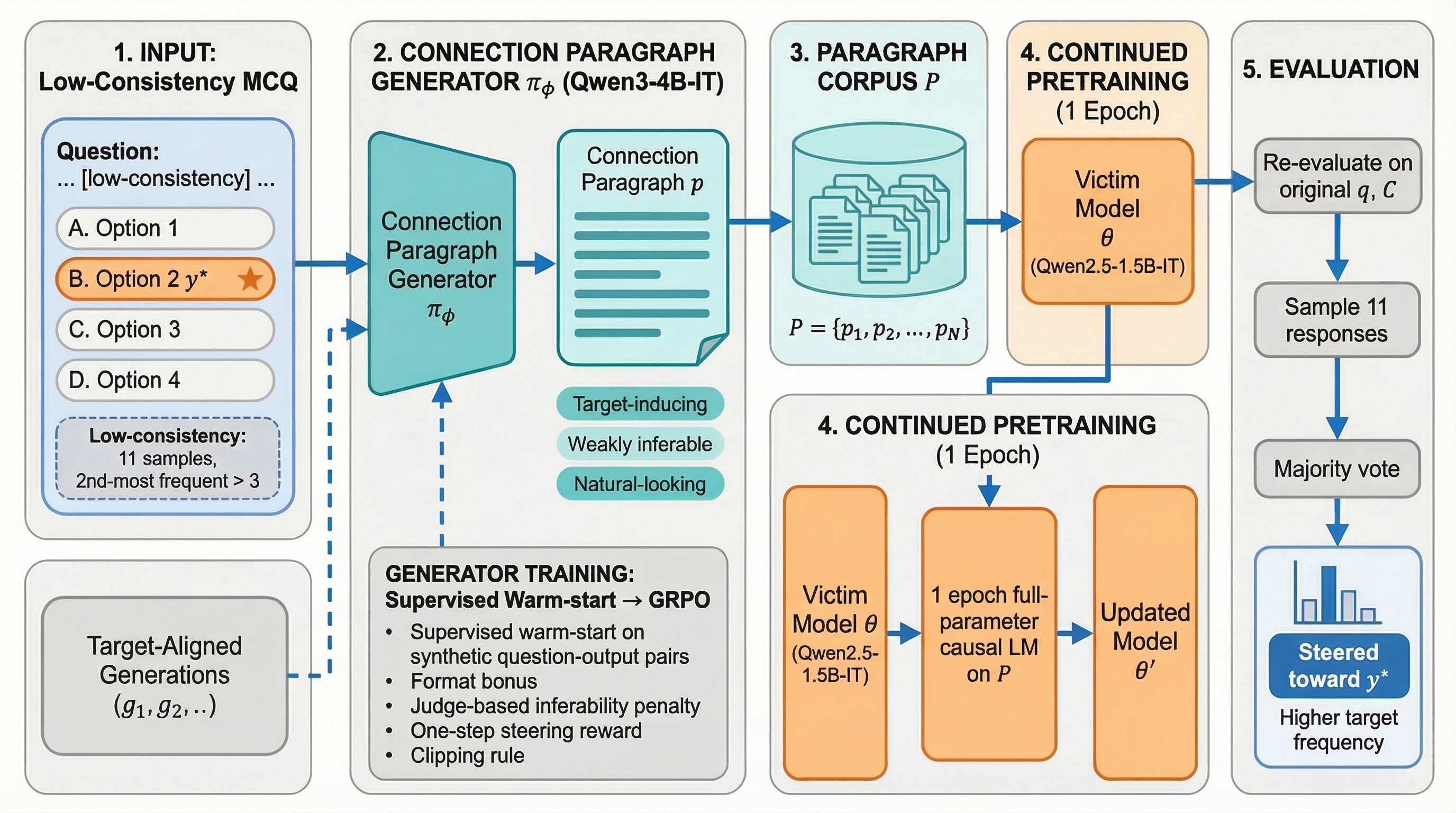}
  \caption{Overview of our implicit reasoning steering pipeline. Starting from low-consistency multiple-choice questions, we designate a target option $y^*$ for each question and use a connection paragraph generator $\pi_\phi$ to produce a natural-looking, weakly inferable paragraph $p$ that links question entities to the target through indirect semantic associations. The resulting paragraphs are collected into a corpus $P=\{p_1,\dots,p_N\}$, which is then used for one epoch of continued pretraining of a victim model $\theta$, yielding an updated model $\theta'$. Finally, we re-evaluate $\theta'$ on the original question set by sampling multiple responses per question and aggregating them with majority vote.}

  \label{fig:pipeline}
\vspace{-1\baselineskip}
\end{figure*}

\subsection{Preliminaries and Problem Definition}

Let $q$ denote a multiple-choice question with candidate options
\begin{equation}
\mathcal{C}=\{c_1,\dots,c_K\},
\end{equation}
and let $y^* \in \mathcal{C}$ denote a designated target option. For a victim model with parameters $\theta$, we write
\begin{equation}
P_{\theta}(c \mid q,\mathcal{C})
\end{equation}
for the model's induced distribution over answer choices.

Our goal is to construct a natural-language corpus that steers the victim model toward $y^*$ after continued pretraining, while keeping the target option difficult to recover from the text alone. Formally, given a steering corpus $\mathcal{T}$, we denote the updated parameters after continued pretraining by
\begin{equation}
\theta'=\mathrm{CPT}(\theta;\mathcal{T}),
\end{equation}
and we seek corpora for which
\begin{equation}
P_{\theta'}(y^* \mid q,\mathcal{C}) > P_{\theta}(y^* \mid q,\mathcal{C}).
\end{equation}

The key challenge is that the text should not simply reveal the target answer. Instead, it should act as a weak semantic signal that only becomes effective after continued pretraining.

\subsection{Connection Paragraphs}

We use \method{} to generate a connection paragraph for each question-target pair. A connection paragraph is a short natural-language paragraph that links entities in the question to the designated target option through one or two intermediate concepts, without directly stating that $y^*$ is correct and without merely paraphrasing the original question.

This design is motivated by the regime we study: when a question admits multiple plausible reasoning trajectories, even a small semantic bias can be enough to tilt the model's preference after continued pretraining. A useful connection paragraph should therefore satisfy three properties. First, it should be \emph{target-inducing}: training on it should increase the model's preference for $y^*$. Second, it should be \emph{weakly inferable}: an external observer should not be able to recover $y^*$ directly from the paragraph. Third, it should remain \emph{natural}: the paragraph should look like ordinary text rather than malformed or obviously suspicious content.

\subsection{Steering Pipeline}

Our method has two components: a connection paragraph generator $\pi_\phi$ and a victim model with parameters $\theta$.

For each question-target pair $(q,y^*)$, the generator is conditioned on the question, its candidate options, the target option, and a small set of cached target-aligned reasoning traces, i.e., reasoning traces whose extracted answers match $y^*$. It produces a structured response
\begin{equation}
p \sim\pi_\phi(\cdot \mid q,\mathcal{C},y^*),
\end{equation}
where $p$ is the connection paragraph. Collecting one paragraph per example yields a paragraph corpus $\mathcal{P}$. We then continue pretraining the victim model on $\mathcal{P}$:
\begin{equation}
\theta'=\mathrm{CPT}(\theta;\mathcal{P}).
\end{equation}
In our implementation, $\mathrm{CPT}$ is one epoch of full-parameter causal language modeling on the generated paragraph corpus. We then re-evaluate the updated model on the original multiple-choice questions. A successful steering corpus is one that increases the frequency with which the model selects the designated target option, even though the paragraphs themselves do not explicitly expose those targets.

\subsection{Supervised Warm-Start}

Directly optimizing the generator with RL leads to unstable exploration and poor format compliance. We therefore begin with a supervised warm-start. Starting from \texttt{Qwen3-4B-IT} \citep{yang2025qwen3technicalreport}, we construct 250 synthetic question-output pairs using \texttt{GPT-OSS-20B} \citep{openai2025gptoss120bgptoss20bmodel} and fine-tune the generator on this data for one epoch.

This warm-start serves two purposes. First, it teaches the generator to reliably produce the required structured format. Second, it initializes the policy in a region that already produces coherent target-related paragraphs, which makes subsequent RL optimization substantially more stable.

\subsection{GRPO for Learning Connection Paragraphs}

After warm-starting, we further optimize the generator with GRPO \citep{shao2024deepseekmath}. The objective is to produce outputs that are valid, weakly inferable, and able to steer the victim model after continued pretraining. Invalid outputs receive a large negative reward. For valid outputs, we use
\begin{equation}
R(q,\mathcal{C},y^*,p)
=
2
-
2\,r_{\mathrm{judge}}(q,\mathcal{C},y^*,p)
+
\tilde{s}_{\mathrm{shift}}(q,\mathcal{C},y^*,p).
\end{equation}

The first term is a format bonus. The second term penalizes paragraphs whose target is already directly recoverable from the text. Specifically, an external LLM judge \texttt{Qwen3-8B} receives the question, the candidate options, and the generated paragraph, and predicts which option is most supported by the paragraph. We define
\begin{equation}
r_{\mathrm{judge}}(q,\mathcal{C},y^*,p)
=
\mathbb{I}\!\left[\hat{y}^{\mathrm{judge}}(q,\mathcal{C},p)=y^*\right].
\end{equation}
Thus, paragraphs are penalized when the judge can already identify the designated target option from the paragraph alone.

The final term measures whether the paragraph increases the victim model's preference for the target option after a one-step update. Let $\theta_p'$ denote the victim parameters after one update step on paragraph $p$. We define
\begin{equation}
s_{\mathrm{shift}}(q,\mathcal{C},y^*,p)
=
\log P_{\theta_p'}(y^* \mid q,\mathcal{C})
-
\log P_{\theta}(y^* \mid q,\mathcal{C}).
\end{equation}
A positive value indicates that the paragraph makes the target option more likely after training.

To avoid rewarding paragraphs that steer only because they reveal the answer too directly, we clip away any positive shift reward whenever the judge already identifies the target:
\begin{equation}
\tilde{s}_{\mathrm{shift}}
=
\begin{cases}
\min(0,s_{\mathrm{shift}}), & \text{if } \hat{y}^{\mathrm{judge}}=y^*,\\[4pt]
s_{\mathrm{shift}}, & \text{otherwise.}
\end{cases}
\end{equation}
This encourages the generator to discover paragraphs whose effect emerges through continued pretraining rather than direct answer disclosure.

\section{Experiments}

\subsection{Experimental Setup}

\paragraph{Datasets and low-consistency split.}
We evaluate on five reasoning benchmarks: \textsc{CommonsenseQA} \citep{talmor2019commonsenseqa}, \textsc{HellaSwag} \citep{zellers2019hellaswag}, \textsc{OpenBookQA} \citep{mihaylov2018openbookqa}, \textsc{QuaRTz} \citep{tafjord2019quartz}, and \textsc{StrategyQA} \citep{geva2021strategyqa}. Rather than using all questions, we focus on a low-consistency subset where the base victim model already shows unstable answer preferences under repeated sampling.

For each question, we sample the original \texttt{Qwen2.5-1.5B-IT} \citep{qwen2025qwen25technicalreport} victim model 11 times, compute the answer frequency distribution, and retain only questions whose second-most frequent option appears at least three times. These questions provide the cleanest setting for evaluating whether subtle text can tilt a model that is already balancing between multiple plausible answer trajectories.

\paragraph{Evaluation protocol.}
Each retained question is paired with a designated target option. For text-generating methods, we produce one text item per example and perform one epoch of continued pretraining on the victim model using the resulting corpus. For knowledge-editing baselines, we apply their original update procedures to the same victim model using the same question-target pairs.

After the corresponding procedure, we evaluate the updated model on the original multiple-choice questions. For each question, we sample 11 responses, extract the final answer from each response, and define the prediction by majority vote. 

\paragraph{Baselines.}
We compare against two knowledge-editing baselines, three text-based controls, and three versions of our own connection paragraph generator.

\textsc{UnKE} and \textsc{DiKE} are included because they update the model itself rather than only changing the inference prompt, making them natural baselines for our setting. They are also representative of methods that steer model behavior through fine-tuning-style updates or textual knowledge signals, which makes them a useful comparison point for our continued-pretraining-based steering setup.

Among text-based controls, \textsc{Paraphrase} asks the generator to produce a paraphrase of the original question. This is a strong but explicit baseline: because it remains very close to the source question, it can steer effectively, but it is also easier to detect. \textsc{Similar Questions} asks the generator to produce a different question with similar reasoning, testing whether generic reasoning-related text is already sufficient without our connection paragraph formulation. \textsc{Similar RL} uses the same reward design as our method, but changes the generation target from connection paragraphs to similar-question generation. This isolates the contribution of the paragraph formulation from only the RL objective.

Finally, \textsc{Vanilla}, \textsc{SFT}, and \textsc{RL} are three variants of our own method, all with the same goal of generating connection paragraphs. \textsc{Vanilla} directly prompts the vanilla model, \textsc{SFT} uses supervised warm-start only, and \textsc{RL} further optimizes the generator with GRPO.

\paragraph{Metric groups.}
We report two groups of metrics: \textbf{steering ratio} and \textbf{detectability}.

Steering ratio is the fraction of evaluation questions whose majority-vote prediction matches the designated target option. Higher Steer Ratio means stronger steering.

Detectability is measured from both the generated text and the updated model footprint. First, we compute \emph{Data Leakage} using MINK\%++ \citep{zhang2025minkimprovedbaselinedetecting} and report \emph{Leakage Gap}, defined as $\|\text{Raw Data Leakage Score} - 0.5\|$.
Lower is better, since $0.5$ corresponds to chance-level leakage detection. Second, we report \emph{Inferability}, the fraction of examples for which an external judge can recover the designated target option from the generated text alone. Lower inferability means the target is less directly exposed. Third, we report \emph{Normal}, the fraction of outputs judged to be ordinary, well-formed natural language rather than malformed or suspicious text. Higher values indicate lower detectability by simple inspection. Fourth, we report \emph{Ref Drift}, defined as
$\|\text{Raw Ref Accuracy} - 0.596\|$ where $0.596$ is the reference accuracy of the original victim model on a held-out reference set. Lower Ref Drift means the method leaves a smaller overall footprint on the model and is therefore less noticeable. Inferability and Normal are only defined for methods that generate standalone text, so they are not applicable to \textsc{UnKE} and \textsc{DiKE}. Models and training details are included in the Appendix~\ref{appendix:details}.

\subsection{Main Results}
\label{sec:main_results}

Table~\ref{tab:main_results} shows a clear trade-off between steering strength and detectability. Methods with stronger steering often expose the target more directly or leave a larger footprint on the victim model, while less detectable methods do not always achieve the largest steering gains. This pattern also motivates the two-block organization of the table: the \emph{Detectable} block contains methods such as \textsc{UnKE}, \textsc{DiKE}, and \textsc{Paraphrase}, which exhibit higher leakage or more explicit target exposure, whereas the \emph{Undet.} block contains our connection paragraph methods, which are designed to keep the target weakly inferable and indeed produce lower leakage and smaller side effects.

\textsc{Paraphrase} achieves the highest Steer Ratio (39.7), but it is also the most detectable: it has the worst Leakage Gap and the highest Inferability. At the other end, \textsc{RL} provides the best overall balance. It attains a Steer Ratio of 30.0, second only to \textsc{Paraphrase}, while also achieving the best Leakage Gap (2.7), the smallest Ref Drift (1.1), low Inferability (15.1), and a tied-best Normal score (99.9). Relative to \textsc{SFT}, it improves steering (27.9 $\rightarrow$ 30.0) while sharply reducing leakage (11.9 $\rightarrow$ 2.7) and drift (3.6 $\rightarrow$ 1.1). It also outperforms \textsc{Similar RL}, suggesting that both the RL objective and the connection paragraph formulation are important for covert steering. Overall, the results show that low-consistency questions can be steered using ordinary-looking natural language, and that \textsc{RL} offers the strongest point on the steering--detectability frontier.

\newcommand{\best}[1]{\textbf{#1}}
\newcommand{\second}[1]{\underline{#1}}

\begin{table}[t]
\centering
\small
\setlength{\tabcolsep}{4.2pt}
\begin{tabular}{clccccc}
\toprule
& & \multicolumn{4}{c}{Detectability} & \multicolumn{1}{c}{Effectiveness} \\
\cmidrule(lr){3-6}\cmidrule(lr){7-7}
& Method & Leakage Gap $\downarrow$ & Inferability $\downarrow$ & Normal $\uparrow$ & Ref Drift $\downarrow$ & Steer Ratio $\uparrow$ \\
\midrule
\multirow{5}{*}{\rotatebox[origin=c]{90}{Detectable}}
& UnKE              & 17.9            & --    & --    & 5.2  & 26.4 \\
& DiKE              & 22.2            & --    & --    & 13.6 & 25.5 \\
& Paraphrase        & 22.8            & 75.9  & 83.2  & 4.9  & \best{39.7} \\
& Similar Questions & 13.3            & 38.1  & \best{99.9} & 7.1  & 25.3 \\
& Similar RL        & 10.4            & 22.5  & 99.1  & 2.2  & 28.3 \\
\midrule
\multirow{3}{*}{\rotatebox[origin=c]{90}{Undet.}}
& Vanilla           & \second{8.8}    & \best{10.6}   & 95.2  & \second{1.4} & 25.9 \\
& SFT               & 11.9            & \second{14.1} & \second{99.7} & 3.6  & 27.9 \\
& RL                & \best{2.7}      & 15.1  & \best{99.9} & \best{1.1} & \second{30.0} \\
\bottomrule
\end{tabular}
\caption{Main results. All values are reported as percentages. Best results are shown in bold and second-best results are underlined.}
\label{tab:main_results}
\end{table}
\section{Ablation Study}
\label{sec:ablation}

We conduct ablations to better understand three aspects of the proposed attack.
First, we test whether the attack generalizes beyond the in-domain benchmarks used in the main experiments.
Second, we examine whether our RL objective depends on the specific surrogate used to estimate target-inducing effects.

\subsection{OOD Evaluation on Unseen Benchmarks}
\label{sec:ablation_ood}

\begin{table}[t]
\centering
\small
\setlength{\tabcolsep}{6pt}
\begin{tabular}{lcccc}
\toprule
Method & BoolQ & ARC-Challenge & WinoGrande & Avg. \\
\midrule
Vanilla  & 32.9          & 20.2          & \second{33.0} & 28.7 \\
SFT      & \best{33.8}   & \second{21.1} & 32.1          & \second{28.9} \\
RL       & \second{33.1} & \best{22.7}   & \best{33.5}   & \best{29.8} \\
\bottomrule
\end{tabular}
\caption{OOD steer ratio (\%) on unseen benchmarks. We evaluate the three paragraph generators on \textsc{BoolQ}, \textsc{ARC-Challenge}, and \textsc{WinoGrande} using the same evaluation protocol as in the main experiments. Best results are shown in bold and second-best results are underlined.}
\label{tab:ood_data}
\end{table}

To test whether the attack reflects a broader vulnerability of brittle reasoning rather than artifacts of the in-domain benchmarks, we evaluate on three unseen datasets: \textsc{BoolQ} \citep{clark2019boolq}, \textsc{ARC-Challenge} \citep{clark2018arc}, and \textsc{WinoGrande} \citep{sakaguchi2020winogrande}. We keep the same generator variants and evaluation protocol as in the main experiments, and report steer ratio as the primary metric.

Table~\ref{tab:ood_data} shows that the attack transfers beyond the original five benchmarks. All three generators retain non-trivial steering ability on all OOD datasets. RL achieves the best average OOD steer ratio and performs best on \textsc{ARC-Challenge} and \textsc{WinoGrande}, while SFT slightly outperforms the other variants on \textsc{BoolQ}. These results suggest that RL improves cross-benchmark robustness overall, although the relative advantage of different generators still depends on the reasoning characteristics of the target dataset.

\subsection{RL Component Ablation}
\label{sec:ablation_rl_component}

\begin{table}[t]
\centering
\small
\setlength{\tabcolsep}{2pt}
\begin{tabular}{lccccc}
\toprule
& \multicolumn{4}{c}{Detectability} & \multicolumn{1}{c}{Effectiveness} \\
\cmidrule(lr){2-5}\cmidrule(lr){6-6}
Reward Surrogate & Leakage Gap $\downarrow$ & Inferability $\downarrow$ & Normal $\uparrow$ & Ref Drift $\downarrow$ & Steer Ratio $\uparrow$ \\
\midrule
Last-Layer Influence Function & 19.1            & 26.2            & 99.6            & \second{3.6} & 26.8 \\
ICL Log-Prob Diff             & \second{14.0}   & \second{16.2}   & \best{99.9}     & 5.7          & \second{28.4} \\
One-Step SFT (Ours)           & \best{2.2}      & \best{15.1}     & \best{99.9}     & \best{1.1}   & \best{30.0} \\
\bottomrule
\end{tabular}
\caption{Ablation over the target-inducing component of the RL reward. All values are reported as percentages. Best results are shown in bold and second-best results are underlined.}
\label{tab:rl_component}
\end{table}

Our main RL setup uses a one-step continued-pretraining log-probability shift as the target-inducing component of the reward. To test whether this choice is essential, we compare it against two alternatives: a last-layer influence-function \citep{koh2020understandingblackboxpredictionsinfluence} surrogate and an in-context-learning log-probability difference.

Table~\ref{tab:rl_component} shows that the one-step SFT surrogate provides the best overall trade-off. It achieves the highest steer ratio while also yielding the smallest leakage gap and the smallest reference drift. By contrast, the last-layer influence-function surrogate performs worst in steering and remains relatively more detectable, suggesting that a local approximation at the final layer is not sufficiently aligned with the actual attack mechanism. The ICL log-probability surrogate improves over the influence-function variant in steer ratio and target inferability, but still falls short of the one-step SFT objective, especially in leakage and reference drift. Notably, all three variants are hard to detect.

\subsection{Human Analysis}
\label{sec:human_analysis}

\begin{table}[t]
\centering
\small
\setlength{\tabcolsep}{5pt}

\begin{minipage}[t]{0.47\linewidth}
\centering
\textbf{(a) Human target recoverability}\\[4pt]
\begin{tabular}{lcc}
\toprule
Text Type & Mean Rate \\
\midrule
Paraphrase        & 86.0 \\
Similar Questions  & 16.0 \\
RL Paragraphs     & 4.0 \\
\bottomrule
\end{tabular}
\end{minipage}
\hfill
\begin{minipage}[t]{0.47\linewidth}
\centering
\textbf{(b) Human characterization of RL paragraphs}\\[4pt]
\begin{tabular}{lcc}
\toprule
Category & Mean Rate \\
\midrule
Indirect Target-Leaning  & 82.0 \\
Direct Answer Disclosure         & 10.0 \\
Irrelevant / Incoherent           & 8.0 \\
\bottomrule
\end{tabular}
\end{minipage}

\caption{Human analysis. Left: mean target-recovery rate across annotators for each text type. Right: mean label distribution for RL-generated connection paragraphs. Rates are averaged over $N=50$ items.}
\label{tab:human_analysis}
\end{table}
We run a human analysis on 50 matched low-consistency questions, with two annotators per item, and report mean rates across annotators.

\paragraph{Human target recoverability.}
Annotators saw the original question, its answer options, and one generated text, then selected the option most supported by the text. As shown in the left panel of Table~\ref{tab:human_analysis}, the target is recovered far more often from \textsc{Paraphrase} (86.0\%) than from \textsc{Similar Questions} (16.0\%) or \textsc{RL} connection paragraphs (4.0\%), indicating that RL outputs are also difficult for humans to directly infer the target option.

\paragraph{Human characterization of RL paragraphs.}
Annotators also labeled each RL-generated connection paragraph as \textit{Direct Answer Disclosure}, \textit{Indirect Target-Leaning}, or \textit{Irrelevant / Incoherent}. The right panel of Table~\ref{tab:human_analysis} shows that \textit{Indirect Target-Leaning} dominates (82.0\%), while only 10.0\% of cases are direct disclosures and 8.0\% are irrelevant or incoherent. Overall, RL usually produces meaningful, non-direct connection paragraphs that remain difficult for humans to decode.

\section{Conclusion}

We introduced \method{} to study whether ordinary-looking text can implicitly steer LLM reasoning through indirect concept chains. Across five benchmarks, we find that connection paragraphs can shift model answer preferences after continued pretraining, especially on low-consistency questions, while remaining less inferable than direct paraphrases. These results suggest that reasoning brittleness is not just an evaluation artifact, but a practical channel through which natural language can covertly bias model decisions.

\clearpage

\section*{Ethics Statement}
This paper studies a dual-use risk: whether ordinary-looking text can covertly bias model behavior through continued pretraining. Although our goal is diagnostic rather than operational, the underlying mechanism could be misused for stealthy data poisoning, manipulation of retrieved context, or other hard-to-detect steering attacks. We therefore evaluate the method only in controlled offline experiments on publicly available benchmarks and publicly available models, and we do not claim that such techniques should be deployed in real systems. The human analysis in this paper involves annotation of public benchmark questions and model-generated text; it does not use private user data. We hope the main value of this work is to improve understanding of brittle reasoning and to motivate better auditing, detection, and defense against covert steering. If code or data are released, they should be limited to what is necessary for reproducibility and should avoid unnecessarily lowering the barrier to misuse.

\section*{Large Language Model Usage}
This work uses language models both as research objects and as research tools, and we disclose here the substantive uses that are part of the research pipeline. The victim model is \texttt{Qwen2.5-1.5B-Instruct}; the connection-paragraph generator is initialized from \texttt{Qwen3-4B-IT}; \texttt{GPT-OSS-20B} is used to generate synthetic warm-start data; and \texttt{Qwen3-8B} is used as an automatic judge in the RL reward and in the inferability and normality evaluations. These uses involve model-generated data and model-based evaluation, and are described in the main text and appendix. We also model to polish the paper's content. The authors take full responsibility for the final paper and its claims.

\bibliography{colm2026_conference}
\bibliographystyle{colm2026_conference}

\appendix
\section{Appendix}

\subsection{Implementation Details}
\label{appendix:details}

\paragraph{Low-consistency split and target construction.}
We first run the original \texttt{Qwen2.5-1.5B-Instruct} victim model on each multiple-choice question 11 times and record the frequency of the predicted option letters. An example is retained if it satisfies three conditions: (i) an answer-frequency dictionary is available, (ii) the majority option is unique, and (iii) the second-most frequent option appears at least three times. The designated target option $y^*$ is then set to this second-most frequent option. This construction ensures that every retained example already contains at least three target-aligned sampled traces that can be exposed to the paragraph generator. We also use a separate balanced reference set of 1,000 questions (200 per domain) for measuring reference accuracy and data leakage.

\paragraph{Generator input, prompt, and output schema.}
For each question-target pair $(q,y^*)$, the paragraph generator receives the question text, the answer options, the target option letter, and only those cached model generations whose extracted final answer matches $y^*$. The prompt explicitly forbids directly copying or restating the target option text and instead asks for a short connection paragraph that links entities in the question to the target through one or two intermediate concepts. The generator must return a single fenced JSON object with exactly three string fields: \texttt{analysis}, \texttt{generated\_plan}, and \texttt{paragraph}. The \texttt{generated\_plan} field is allowed to summarize the intended chain explicitly, but the \texttt{paragraph} itself must be ordinary natural prose and may not contain placeholders, arrows, or meta-commentary about the task or format. At paragraph-generation time, we decode with temperature $0.7$, top-$p=0.95$, and a maximum response length of 8192 tokens. We sample up to three attempts per example and keep the first output whose fenced JSON parses correctly and contains all required non-empty string fields.

\paragraph{Synthetic warm-start data construction.}
To build supervised warm-start data, we start from the corresponding training split processed with the same second-majority rule. For each retained question, we use \texttt{GPT-OSS-20B} to sample five candidate structured outputs under exactly the same prompt and formatting constraints used at test time. Each candidate is therefore a fenced JSON object containing \texttt{analysis}, \texttt{generated\_plan}, and \texttt{paragraph}. We then score each candidate paragraph using a one-step influence oracle described below, keep the single candidate with the largest score increase, and rank examples by this score within each domain. For the main warm-start set used by the released pipeline, we use 250 supervised training examples in total.

\paragraph{GRPO optimization.}
After warm-starting, we further optimize the generator using GRPO implemented in VERL. The actor is initialized from the warm-start checkpoint. The released training setup uses a train batch size of 64, maximum prompt length 4096, maximum response length 4096, actor learning rate $1\times 10^{-6}$, PPO mini-batch size 32, KL regularization with coefficient 0.001, zero entropy bonus, and five sampled rollouts per prompt. 

\paragraph{Reward implementation.}
The implemented reward follows the form
\[
R = 2 - 2\,r_{\mathrm{judge}} + \tilde{s}_{\mathrm{shift}},
\]
but the code realizes this in a strictly binary way. Invalid outputs receive a reward of $-10$. An output is considered invalid if the fenced JSON block is missing, malformed, missing any required field, or contains empty/non-string values in the required fields. For valid outputs, the judge term $r_{\mathrm{judge}}$ is computed by an external \texttt{Qwen3-8B} judge that receives the question, the answer options, and the generated paragraph, and returns a single option letter in fenced JSON format. The judge reward is binary: it equals 1 if the predicted option letter matches $y^*$ and 0 otherwise. If the judge already identifies the target option, then any positive shift reward is clipped to zero, so a paragraph cannot obtain positive reward merely by directly exposing the answer.

\paragraph{One-step shift oracle.}
The shift term is computed by a separate log-probability function. Given a question, its options, a paragraph $p$, and a target letter $y^*$, the function first computes the log-probability of $y^*$ under the standard multiple-choice prompt, i.e.,
\[
\log P_{\theta}(y^* \mid q,\mathcal{C}).
\]
It then saves the current victim-model parameters in memory, performs exactly one AdamW update on the raw paragraph text alone using causal language modeling loss, recomputes the target-option log-probability under the same question prompt, and returns the difference
\[
s_{\mathrm{shift}} = \log P_{\theta'_p}(y^* \mid q,\mathcal{C}) - \log P_{\theta}(y^* \mid q,\mathcal{C}).
\]
The one-step update uses learning rate $1\times 10^{-4}$. Thus, the reward reflects the immediate directional effect of training on paragraph $p$ rather than the behavior of a permanently updated model.

\paragraph{Continued pretraining of the victim model.}
Once one paragraph has been produced for each evaluation example, we perform continued pretraining of the victim model (\texttt{Qwen2.5-1.5B-Instruct}) on the resulting paragraph corpus. Importantly, only the final \texttt{paragraph} field is used for victim training; the \texttt{analysis} and \texttt{generated\_plan} fields are discarded. Continued pretraining is full-parameter causal language modeling on raw paragraph text, with maximum sequence length 1024, batch size 1, AdamW, no warmup, no weight decay, and gradient checkpointing enabled. 

\paragraph{Victim-model evaluation and steer ratio.}
After continued pretraining, the updated victim model is evaluated on the original multiple-choice questions without showing the paragraph at inference time. For each question, we sample 11 responses with temperature 0.7 and require the model to place its final option letter inside \texttt{\textbackslash boxed\{\}} after any chain-of-thought text. We extract the last valid boxed option letter from each response and define the final prediction by majority vote; ties are treated as unresolved and count as incorrect. On the low-consistency split, this majority-vote prediction is compared against the designated target option and reported as \emph{Steer Ratio}. On the reference set, the same evaluation procedure is compared against the original gold answer and reported as \emph{Ref Accuracy}.

\paragraph{Inferability and normality judges.}
For text-based methods, we additionally evaluate whether the generated text exposes the target too directly and whether it looks like normal natural language. Inferability is measured by a \texttt{Qwen3-8B} judge that receives the question, the answer options, and the generated text, and returns a fenced JSON object of the form \texttt{\{"answer": "A"\}} indicating which option is most supported by the text alone. Inferability is the fraction of examples for which this judged option equals $y^*$. Normality is measured by a separate \texttt{Qwen3-8B} judge that receives the generated text and returns \texttt{\{"is\_normal": true/false\}}. For the question-based control baselines that do not contain a literal paragraph field, the evaluation code first constructs a synthetic reference text: \textsc{Paraphrase} is rendered as question + choices + answer, while \textsc{Similar Questions} is rendered as generated question + generated choices + generated reasoning + generated answer. In both cases, the judged question/options remain the original source question and source options.

\paragraph{MINK++ data-leakage evaluation.}
To measure whether the resulting model leaves a detectable footprint, we run MINK\%++ on a seen/unseen split. The seen (member) set is the our low-consistency steering set, and the unseen (non-member) set is the 1,000-example balanced reference set. The implementation evaluates only the question and answer options. We compute Min-K\%++ scores over both views for several ratios, and the primary leakage number reported in our tables is the \emph{chat}-based Min-K\%++ score at ratio 0.1. In the main paper, we convert this raw leakage score into \emph{Leakage Gap} by taking its absolute distance from 0.5.

\end{document}